%% file: main.tex
\def\year{2021}\relax
\documentclass[letterpaper]{article}
\usepackage{aaai21} 
\usepackage{times} 
\usepackage{helvet} 
\usepackage{courier} 
\usepackage[hyphens]{url} 
\usepackage{graphicx} 
\usepackage{graphicx} 
\usepackage{natbib} 
\usepackage{caption} 
\input{prestuff.tex}

\usepackage[frozencache]{minted}
\usepackage{amssymb}

\newcommand{\ajoin}{\,\rotatebox[origin=c]{90}{$\lozenge$}\,} 
\newcommand{\system}{\textsf{i-Algebra}\xspace}
\setmintedinline{fontsize=\scriptsize}

\title{i-Algebra: Towards Interactive Interpretability of Deep Neural Networks}

\author{Xinyang Zhang\textsuperscript{$\dagger$} \quad
Ren Pang\textsuperscript{$\dagger$} \quad
Shouling Ji\textsuperscript{$\ddagger$} \quad
Fenglong Ma\textsuperscript{$\dagger$} \quad
Ting Wang\textsuperscript{$\dagger$}\\
}

\affiliations{
\textsuperscript{$\dagger$}Pennsylvania State University, \{xqz5366, rbp5354, fenglong, ting\}@psu.edu\\
\quad \textsuperscript{$\ddagger$}Zhejiang University, sji@zju.edu.cn
}

\begin{document}
\maketitle

\input{abstract.tex}

\input{introduction.tex}

\input{background.tex}
\input{method.tex}
\input{analysis.tex}
\input{evaluation.tex}

\input{literature.tex}
\input{conclusion.tex}

\newpage

\section*{Acknowledgments}
This work is supported by the National Science Foundation under Grant No. 1951729, 1953813, and 1953893. Any opinions, findings, and conclusions or recommendations are those of the authors and do not necessarily reflect the views of the National Science Foundation. Shouling Ji was partly supported by the National Key Research and Development Program of China under No. 2018YFB0804102, NSFC under No. 61772466, U1936215, and U1836202, the Zhejiang Provincial Natural Science Foundation for Distinguished Young Scholars under No. LR19F020003, the Zhejiang Provincial Key R\&D Program under No. 2019C01055, the Ant Financial Research Funding, and the Fundamental Research Funds for the Central Universities (Zhejiang University NGICS Platform). We thank Hua Shen for contributing to implementing and evaluating \system.

\newcommand{\bibpre}{bibs}{\small
\bibliography{\bibpre/aml,\bibpre/debugging,\bibpre/general,\bibpre/ting,\bibpre/optimization,\bibpre/graph,\bibpre/interpretation, \bibpre/interaction}}


\end{document}

%% file: prestuff.tex
\usepackage{epsfig,amsmath,amsfonts,epsfig,multirow,makecell,caption,soul,csquotes,color,wrapfig,subcaption,mathtools,bm,spverbatim,booktabs,tcolorbox,diagbox,todonotes}
\usepackage[e]{esvect}

\usepackage{caption}
\makeatletter
\newif\if@restonecol
\makeatother

\usepackage[boxed, ruled, vlined, linesnumbered]{algorithm2e}
\SetKwRepeat{Do}{do}{while}

\newenvironment{changemargin}[2]{\begin{list}{}{
	\setlength{\topsep}{0pt}\setlength{\leftmargin}{0pt}
	\setlength{\rightmargin}{0pt}
	\setlength{\listparindent}{\parindent}
	\setlength{\itemindent}{\parindent}
	\setlength{\parsep}{0pt plus 1pt}
	\addtolength{\leftmargin}{#1}\addtolength{\rightmargin}{#2}
	}\item}
	{\end{list}}

\newenvironment{mitemize}{
	\begin{changemargin}{-3pt}{-0cm}
	\vspace{-10pt}
	\hspace{-5pt}
	\begin{itemize}
	\setlength{\itemsep}{3pt}}
	{\end{itemize}
	\vspace{2pt}
	\end{changemargin}}

\newcommand{\ssub}[2]{{#1}_{\scaleobj{0.8}{#2}}}

\usepackage{titlesec}
\titlespacing\section{0pt}{6pt}{4pt}
\titlespacing\subsection{0pt}{3pt}{2pt}
\titlespacing\subsubsection{0pt}{2pt}{1pt}
\titleformat{\subsection}{\large\bfseries}{\thesubsection}{1em}{}

\usepackage[first=0,last=9]{lcg}
\usepackage{colortbl}
\definecolor{Gray}{gray}{0.8}

\usepackage{hyperref}

\newcommand{\msec}[1]{\S\,\ref{#1}}
\newcommand{\mref}[1]{\,\ref{#1}}
\newcommand{\meq}[1]{Eqn\,(\ref{#1})}
\newcommand{\mcite}[1]{\,\cite{#1}}

\usepackage{scalerel}[2016/12/29]

\makeatletter
\providecommand{\leadsfrom}{%
  \mathrel{\mathpalette\reflect@squig\relax}%
}
\newcommand{\reflect@squig}[2]{%
  \reflectbox{$\m@th#1\leadsto$}%
}
\makeatother

\newcommand{\bx}{\ssub{x}{\circ}}

\newcommand{\dnn}{DNN\xspace}
\newcommand{\dnns}{DNNs\xspace}

\def\eqref#1{equation~\ref{#1}}

\def\1{\bm{1}}

\DeclareMathAlphabet{\mathsfit}{\encodingdefault}{\sfdefault}{m}{sl}
\SetMathAlphabet{\mathsfit}{bold}{\encodingdefault}{\sfdefault}{bx}{n}

\def\gC{{\mathcal{C}}}

\def\gT{{\mathcal{T}}}

\def\gX{{\mathcal{X}}}



%% file: abstract.tex
\begin{abstract}
    Providing explanations for deep neural networks (DNNs) is essential for their use in domains wherein the interpretability of decisions is a critical prerequisite. Despite the plethora of work on interpreting DNNs, most existing solutions offer interpretability in an ad hoc, one-shot, and static manner, without accounting for the perception, understanding, or response of end-users, resulting in their poor usability in practice.

    In this paper, we argue that DNN interpretability should be implemented as the interactions between users and models. We present \system, a first-of-its-kind interactive framework for interpreting DNNs. At its core is a library of atomic, composable operators, which explain model behaviors at varying input granularity, during different inference stages, and from distinct interpretation perspectives. Leveraging a declarative query language, users are enabled to build various analysis tools (e.g., ``drill-down'', ``comparative'', ``what-if'' analysis) via flexibly composing such operators. We prototype \system and conduct user studies in a set of representative analysis tasks, including inspecting adversarial inputs, resolving model inconsistency, and cleansing contaminated data, all demonstrating its promising usability.
\end{abstract}

%% file: introduction.tex
\section{Introduction}
\label{sec:intro}

The recent advances in deep learning have led to breakthroughs in a number of long-standing artificial intelligence tasks, enabling use cases previously considered strictly experimental. Yet, the state-of-the-art performance of deep neural networks (\dnns) is often achieved at the cost of their {\em interpretability}: it is challenging to understand how a \dnn arrives at a particular decision, due to its high non-linearity and nested structure\mcite{Goodfellow-et-al-2016}. This is a major drawback for domains wherein the interpretability of decisions is a critical prerequisite.

\begin{figure}
\centering
\includegraphics[width=50mm]{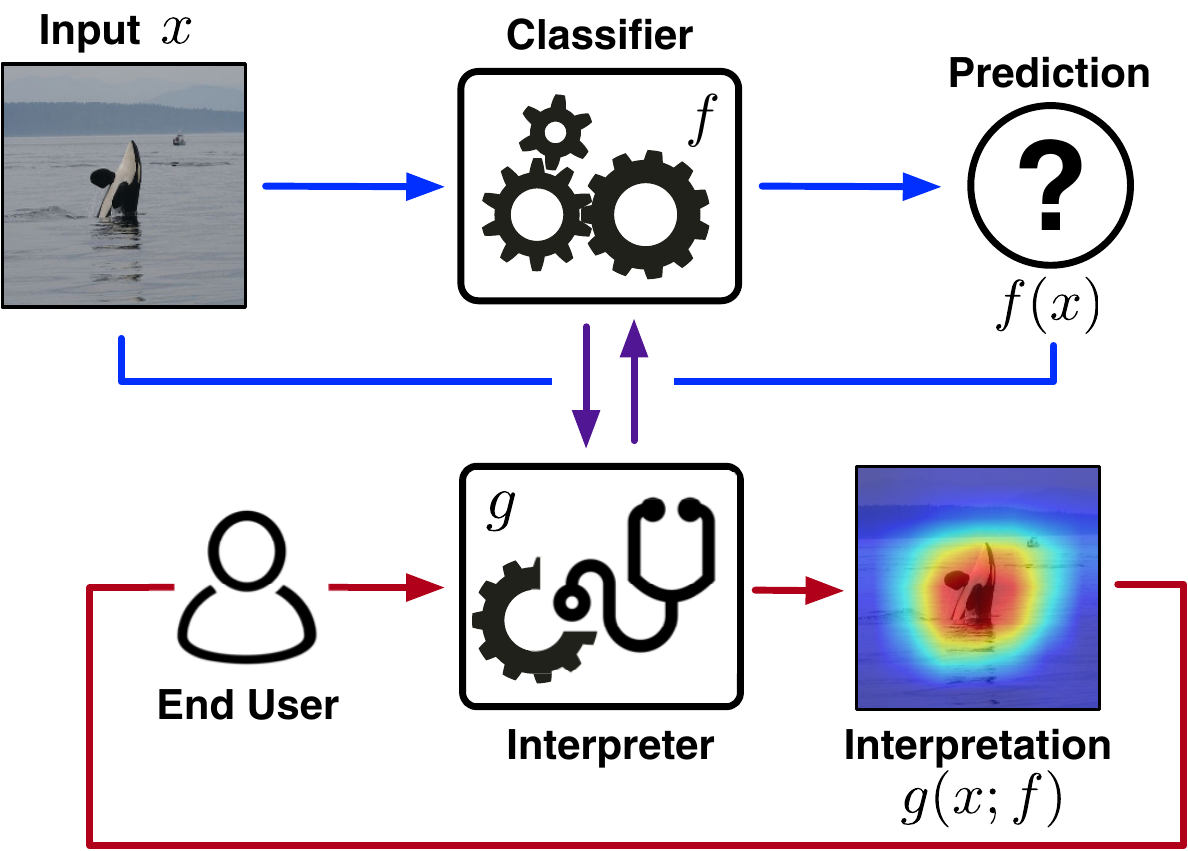}
\caption{Interactive interpretation of DNNs. \label{fig:idls}}
\end{figure}

A flurry of interpretation methods\mcite{Sundararajan:2017:icml,Dabkowski:nips:2017,fong:mask,Zhang:2018:cvpr} have since been proposed to help understand the inner workings of \dnns. For example, in Figure\mref{fig:idls}, the attribution map highlights the most informative features of input $x$ with respect to model prediction $f(x)$. A \dnn (classifier), coupled with an interpretation model (interpreter), forms an interpretable deep learning system (IDLS), which is believed to improve the model trustworthiness\mcite{tao:nips:2018,Guo:2018:ccs}.

Yet, despite the enhanced interpretability, today's IDLSes are still far from being practically useful. In particular, most IDLSes provide interpretability in an ad hoc, single-shot, and static manner, without accounting for the perception, understanding, and response of the users, resulting in their poor usability in practice. For instance, most IDLSes generate a static saliency map to highlight the most informative features of a given input; however, in concrete analysis tasks, the users often desire to know more, for instance,
\begin{mitemize}
\setlength\itemsep{0.5pt}
\item How does the feature importance change if some other features are present/absent?
\item How does the feature importance evolve over different stages of the DNN model?
\item What are the common features of two inputs that lead to their similar predictions?
\item What are the discriminative features of two inputs that result in their different predictions?
\end{mitemize}
Moreover, the answer to one question may trigger followup questions from the user, giving rise to an interactive process. Unfortunately, the existing IDLSes, limited by their non-interactive designs, fail to provide interpretability tailored to the needs of individual users.

\vspace{2pt}
\textbf{Our Work --}
To bridge the striking gap, we present \system, a first-of-its-kind interactive framework for interpreting DNN models, which allows non-expert users to easily explore a variety of interpretation operations and perform interactive analyses. The overall design goal of \system is to implement the DNN interpretability as the interactions between the user and the model.

\begin{figure}
\centering
\includegraphics[width=70mm]{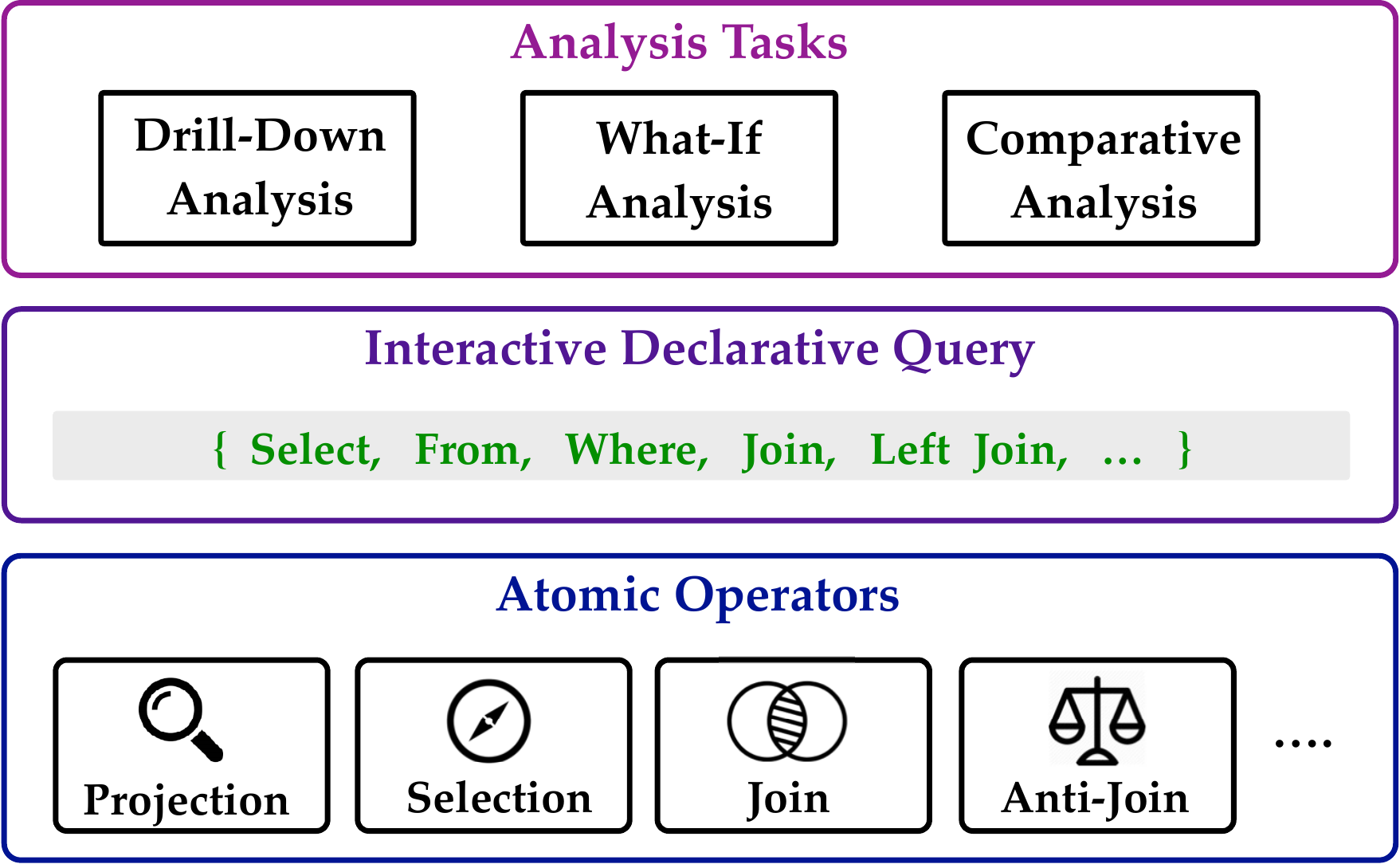}
\caption{\small A framework of interactive interpretation of DNNs.
\label{fig:stack}}
\end{figure}

Specifically, to accommodate a range of user preferences for interactive modes with respect to different DNN models and analysis tasks, we design an expressive algebraic framework, as shown in Figure\mref{fig:stack}. Its fundamental building blocks are a library of atomic operators, which essentially produce DNN interpretability at varying input granularity, during different model stages, and from complementary inference perspectives. On top of this library, we define a SQL-like declarative query language, which allows users to flexibly compose the atomic operators and construct a variety of analysis tasks (e.g., ``drill-down,'' ``what-if,'' ``comparative'' analyses). As a concrete example, given two inputs $x$ and $x'$, the query below compares their interpretation at the $l$-th layer of the DNN $f$ and finds the most discriminative features of $x$ and $x'$ with respect to their predictions.
\vspace{0.5pt}
\begin{minted}[mathescape,
escapeinside=||,
fontsize=\scriptsize,
]{sql}
select l from f(x) left join (select l from f(x|'|))
\end{minted}

\vspace{0.5pt}
We prototype \system and evaluate its usability in three representative tasks: resolving model inconsistency, inspecting adversarial inputs, and cleansing contaminated data. The studies conducted on Amazon MTurk show that compared with the conventional interpretability paradigm, \system significantly improves the analysts' performance. For example, in the task of resolving model inconsistency, we observe over 30\% increase in the analysts' accuracy of identifying correct predictions and over 29\% decrease in their task execution time; in the task of identifying adversarial inputs, \system improves the analysts' overall accuracy by 26\%; while in the task of cleansing poisoning data, \system helps the analysts' detecting over 60\% of the data points misclassified by the automated tool.

\vspace{2pt}
\textbf{Our Contributions --}
To our best knowledge, \system represents the first framework for interactive interpretation of DNNs. Our contributions are summarized as follows.
\begin{mitemize}
\setlength\itemsep{0.5pt}
\item We promote a new paradigm for interactive interpretation of DNN behaviors, which accounts for the perception, understanding, and responses of end-users.
\item We realize this paradigm with an expressive algebraic framework built upon a library of atomic interpretation operators, which can be flexibly composed to construct various analysis tasks.
\item We prototype \system and empirically evaluate it in three representative analysis tasks, all showing its promising usability in practice.
\end{mitemize}

%% file: background.tex
\section{Background and Overview}
\label{sec:background}

\subsection{DNN Interpretation}

We primarily focus on predictive tasks (e.g., image classification): a DNN $f$ represents a function $f: \gX \rightarrow \gC$, which assigns a given input $x \in \gX$ to one of a set of predefined classes $\gC$. We mainly consider post-hoc, instance-level interpretation, which explains the causal relationship between input $x$ and model prediction $f(x)$. Such interpretations are commonly given in the form of {\em attribution maps}. As shown in Figure\mref{fig:idls}, the interpreter $g$ generates an attribution map $m = g(x; f)$, with its $i$-th element $m[i]$ quantifying the importance of $x$'s $i$-th feature $x[i]$ with respect to $f(x)$.

\subsection{Overview of \system}

Despite the rich collection of interpretation models, they are used in an ad hoc and static manner within most existing IDLSes, resulting in their poor usability in practice\mcite{Zhang:2020:sec}. To address this, it is essential to account for the perception, understanding, and response of end-users. We achieve this by developing \system, an interactive framework that allows users to easily analyze DNN's behavior through the lens of interpretation.

\vspace{1pt}
\textbf{Mechanisms --} \system is built upon a library of composable atomic operators, which provides interpretability at different input granularities (e.g., within a user-selected window), during different model stages (e.g., at a specific DNN layer), and from different inference perspectives (e.g., finding discriminative features). Note that we only define the functionality of these operators, while their implementation can be flexibly based on concrete interpretation models. 

\vspace{1pt}
\textbf{Interfaces --}
On top of this library, we define an SQL-like declarative query language to allow users to flexibly compose the operators to construct various analysis tasks (e.g., ``drill-down,'' ``what-if,'' ``comparative'' analysis), which accommodates the diversity of analytical needs from different users and circumvents the ``one-size-fits-all'' challenge.

%% file: method.tex
\section{An Interpretation Algebra}
\label{sec:algebra}

We begin by describing the library of atomic operators. Note that the operators can inherently be extended and all the operators are defined in a declarative manner, independent of their concrete implementation.

\subsection{Atomic Operators}

At the core of \system is a library of atomic operators, including {\em identity}, {\em projection}, {\em selection}, {\em join}, and {\em anti-join}. We exemplify with the Shapley value framework \mcite{Ancona:2019:icml,Chen:2019a:iclr} to illustrate one possible implementation of \system, which can also be implemented based on other interpretation models.

\vspace{2pt}
\textbf{Identity --} The {\em identity} operator represents the basic interpretation $\phi(x; \bar{x}, f)$, which generates the interpretation of a given input $x$ with respect to the DNN $f$ and a baseline input $\bar{x}$ (e.g., an all-zero vector).\footnote{When the context is clear, we omit $\bar{x}$ and $f$ in the notation.} Conceptually, the identity operator computes the expected contribution of $x$'s each feature to the prediction $f(x)$.
Within the Shapley framework, with $x$ as a $d$-dimensional vector ($x \in \mathbb{R}^d$) and $I_k$ as a $k$-sized subset of $I = \{1, 2, \ldots, d\}$, we define a $d$-dimensional vector $\ssub{x}{I_k}$, with its $i$-th dimension defined as:
\begin{equation}
[\ssub{x}{I_k}]_i = \left\{
\begin{array}{ll}
x_i & (i \in I_k) \\
\bar{x}_i & (i \not\in I_k)
\end{array}
\right.
\end{equation}

Intuitively, $\ssub{x}{I_k}$ substitutes $\bar{x}$ with $x$ along the dimensions of $I_k$. Then the attribution map is calculated as:
\begin{align}
\label{eq:basic}
[\phi(x)]_i = \frac{1}{d} \sum_{k=0}^{d-1} \ssub{\mathbb{E}}{I_k}[f(\ssub{x}{I_k \cup \{i\} }) - f( \ssub{x}{I_k})]
\end{align}
where $I_k$ is randomly sampled from $I \setminus \{i\}$.

\begin{figure}[!ht]
\centering
\includegraphics[width=50mm]{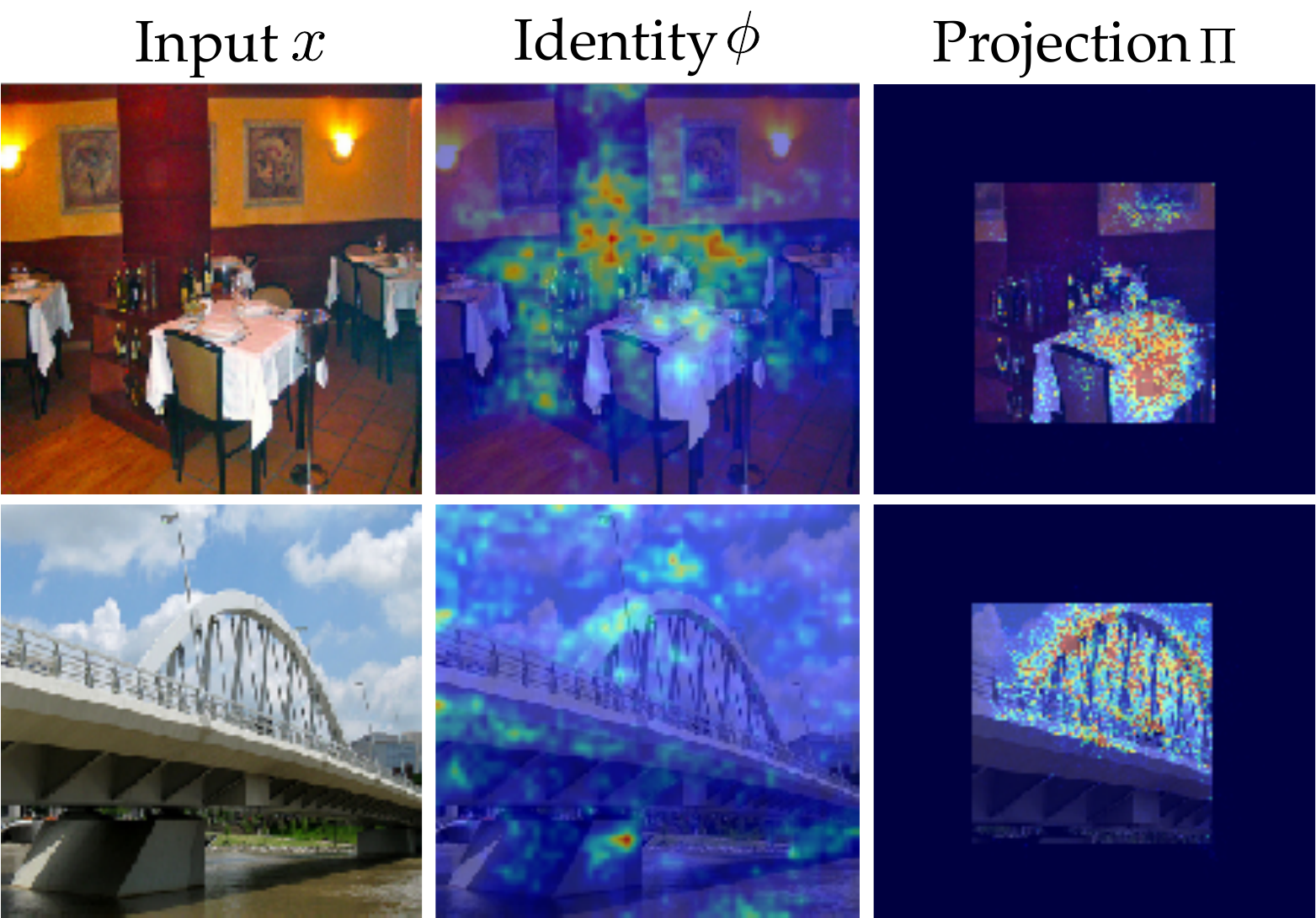}
\caption{Sample inputs and their interpretation under the identity and projection operators (ImageNet and ResNet50). \label{fig:projection}}
\end{figure}

\vspace{2pt}
\textbf{Projection --} While the basic interpretation describes the global importance of $x$'s features with respect to its prediction $f(x)$, the user may wish to estimate the local importance of a subset of features. Intuitively, the global importance approximates the decision boundary in a high dimensional space, while the local importance focuses on a lower-dimensional space, thereby being able to describe the local boundary more precisely.

The {\em projection} operator $\Pi$ allows the user to zoom in a given input. For an input $x$ and a window $w$ (on $x$) selected by the user, $\Pi_w(x)$ generates the local importance of $x$'s features within $w$. To implement it within the Shapley framework, we marginalize $x$'s part outside the window $w$ with the baseline input $\bar{x}$ and compute the marginalized interpretation. Let $w$ corresponds to the set of indices $\{w_1, w_2, \ldots, w_{|w|}\}$ in $I$. To support projection, we define the coalition $\ssub{I}{k}$ as a $k$-sized subset of $\{w_1, w_2, \ldots, w_{|w|}\}$, and redefine the attribution map as: $[\Pi_w(x)]_i =$
\begin{align}
\label{eq:prj}
\left\{
\begin{array}{cl}
\frac{1}{|w|} \sum_{k=0}^{|w|-1} \ssub{\mathbb{E}}{I_k}[f(\ssub{x}{I_k \cup \{i\} }) - f(\ssub{x}{I_k})] & i \in w\\
0 & i \not\in w
\end{array}
\right.
\end{align}

Figure\mref{fig:projection} illustrates a set of sample images and their interpretation under the identity and projection operators (within user-selected windows). Observe that the projection operator highlights how the model's attention shifts if the features out of the window are nullified, which is essential for performing the ``what-if'' analysis.

\vspace{2pt}
\textbf{Selection --} While the basic interpretation shows the static importance of $x$'s features, the user may also be interested in understanding how the feature importance varies dynamically throughout different inference stages. Intuitively, this dynamic importance interpretation captures the shifting of the ``attention'' of DNNs during different stages, which helps the user conduct an in-depth analysis of the inference process\mcite{Karpathy:2016:iclr}.

\begin{figure}[!ht]
\centering
\includegraphics[width=50mm]{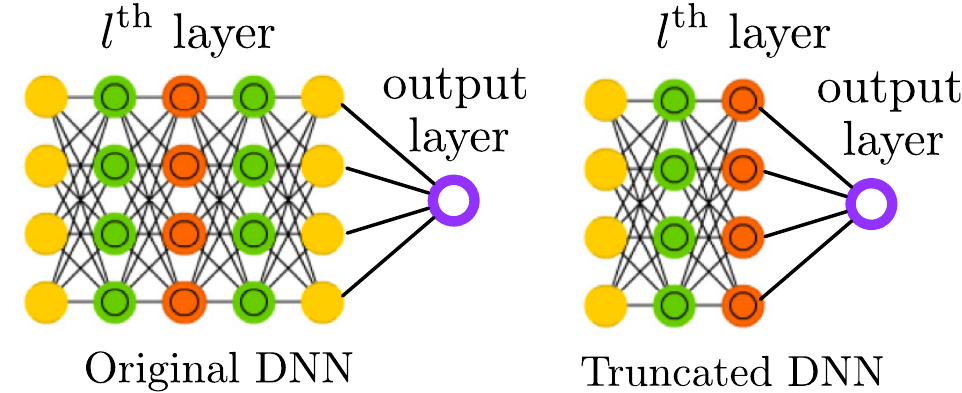}
\caption{Implementation of the selection operator. \label{fig:noperator}}
\end{figure}

The selection operator $\sigma$ allows the user to navigate through different stages of DNNs and investigate the dynamic feature importance. Given an input $x$, a DNN $f$ which consists of $n$ layers $f_{[1:n]}$, and the layer index $i$ selected by the user, $\sigma_i(x)$ generates the interpretation at the $i$-th layer.

One possible implementation of $\sigma_l(x)$ is as follows. We truncate the DNN $f$ at the $l$-th layer, concatenate it with the output layer, and re-train the linear connections between the $l$-th layer and the output layer, as illustrated in Figure\mref{fig:noperator}. Let $f_l$ denote the truncated DNN. Then the selection operator generates a $d$-dimensional map $\sigma_l(x)$ defined as:
\begin{align}
\label{op:selection}
[\sigma_l(x)]_i = [\phi(x; \bar{x}, f_l)]_i
\end{align}
which substitutes $f$ in \meq{eq:basic} with the truncated DNN $f_l$.

\begin{figure}[!ht]
\centering
\includegraphics[width=60mm]{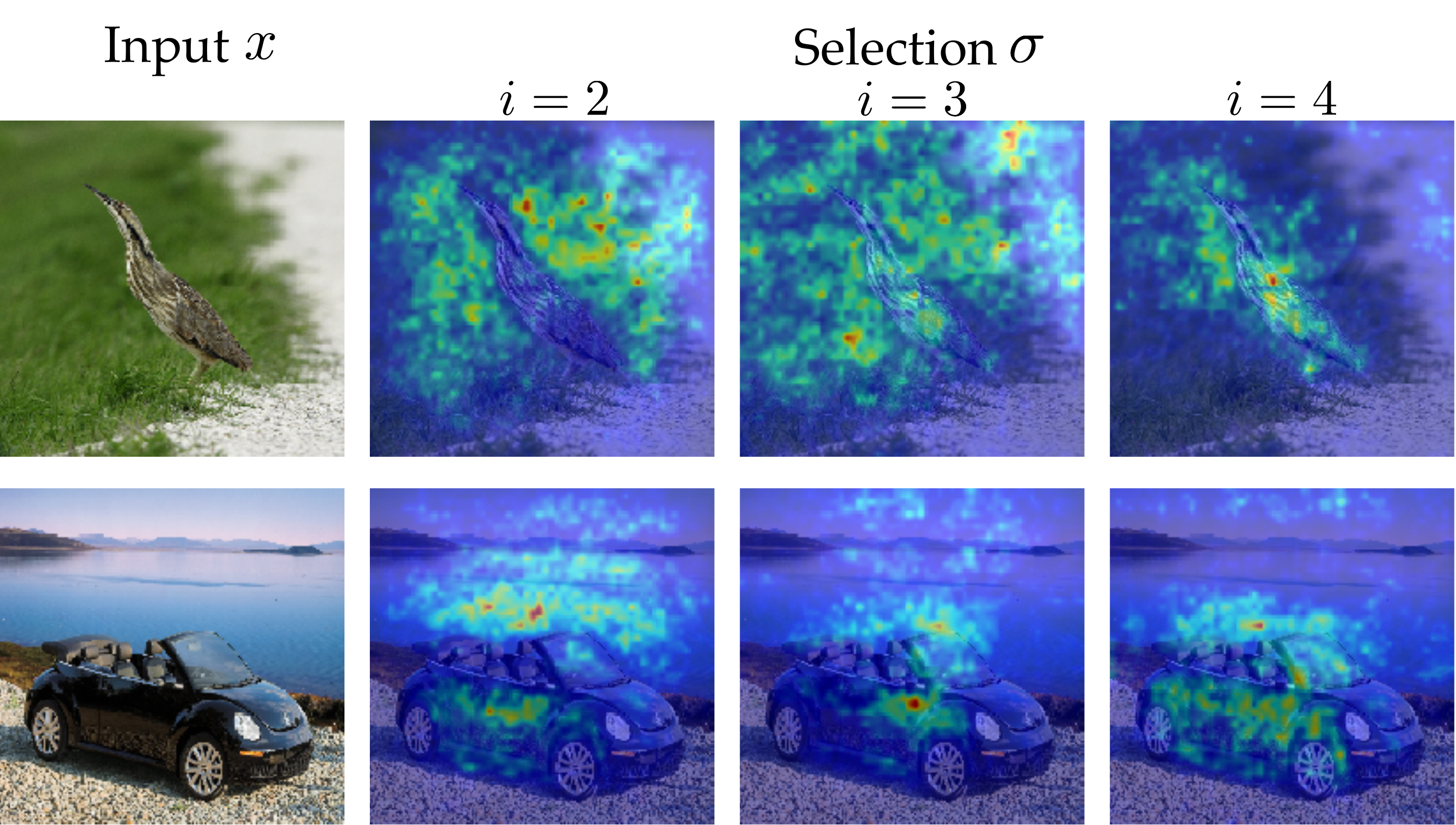}
\caption{Sample inputs and their interpretation under the selection operator (ImageNet and ResNet50). \label{fig:selection}}
\end{figure}

Figure\mref{fig:selection} illustrates a set of sample inputs (from ImageNet) and their attribution maps under the selection operator. Specifically, we select $i = 2, 3, 4$ of the DNN (ResNet50). It is observed that the model's attention gradually focuses on the key objects within each image as $i$ increases.

\vspace{2pt}
\textbf{Join --} There are also scenarios in which the user desires to compare two inputs $x$ and $x'$ from the same class and find the most informative features shared by $x$ and $x'$, from the perspective of the DNN model. The {\em join} of two inputs $x$ and $x'$, denoted by $x \bowtie x'$, compares two inputs and generates the interpretation highlighting the most informative features shared by $x$ and $x'$. Note that the extension of this definition to the case of multiple inputs is straightforward.

Within the Shapley framework, $x \bowtie x'$ can be implemented as the weighted sum of the Shapley values of $x$ and $x'$ (given the weight of $x$'s map as $\epsilon$):
\begin{align}
\label{op:join}
[x \bowtie x']_i = \epsilon \cdot [\phi(x; \bar{x}, f)]_i + (1-\epsilon) \cdot [\phi(x'; \bar{x}, f)]_i
\end{align}

Intuitively, a large value of $[x \bowtie x']_i$ tends to indicate that the $i$-th feature is important for the predictions on both $x$ and $x'$ (with respect to the baseline $\bar{x}$).

\begin{figure}[!ht]
\centering
\includegraphics[width=60mm]{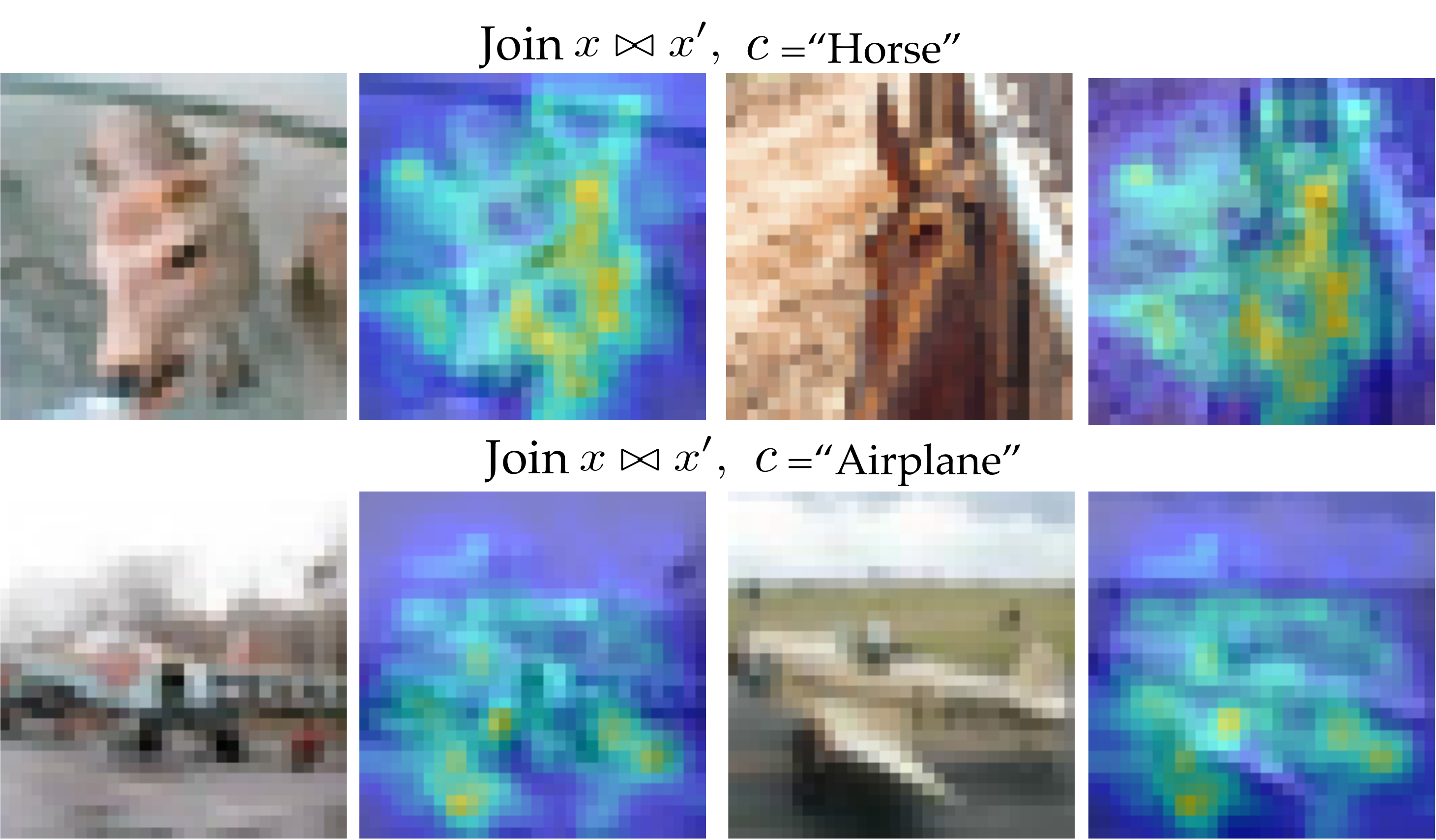}
\caption{Sample inputs and their interpretation under the join operator (CIFAR10 and VGG19). \label{fig:join}}
\end{figure}

Figure\mref{fig:join} illustrates two pairs of sample inputs and their attribution maps under the join operator as $\epsilon=0.5$, which highlight the most important features with respect to their predictions (``horse'' and ``plane'') shared by both inputs.

\vspace{2pt}
\textbf{Anti-Join --} Related to the join operator, the {\em anti-join} of two inputs $x$ and $x'$, denoted by $x \ajoin x'$, compares two inputs $x$ and $x'$ from different classes and highlights their most informative and discriminative features. For instance, in image classification, the user may be interested in finding the most contrastive features of two images that result in their different classifications.

Within the Shapley value framework, the anti-join operator $x \ajoin x'$ can be implemented as the attribution map of $x$ with respect to $x'$ and that of $x'$ with respect to $x$:
\begin{align}
\label{op:ajoin}
[x \ajoin x']_i = ([\phi(x; x', f)]_i, [\phi(x'; x, f)]_i)
\end{align}

It is worth comparing \meq{op:join} and \meq{op:ajoin}: \meq{op:join} compares $x$ (and $x'$) with the baseline $\bar{x}$, highlighting the contribution of each feature of $x$ (and $x'$) with respect to the difference $f(x)- f(\bar{x})$ (and $f(x')- f(\bar{x})$); meanwhile, \meq{op:ajoin} compares $x$ and $x'$, highlighting the contribution of each feature of $x$ (and $x'$) with respect to the difference $f(x)- f(x')$ (and $f(x')- f(x)$).

\begin{figure}[!ht]
\centering
\includegraphics[width=60mm]{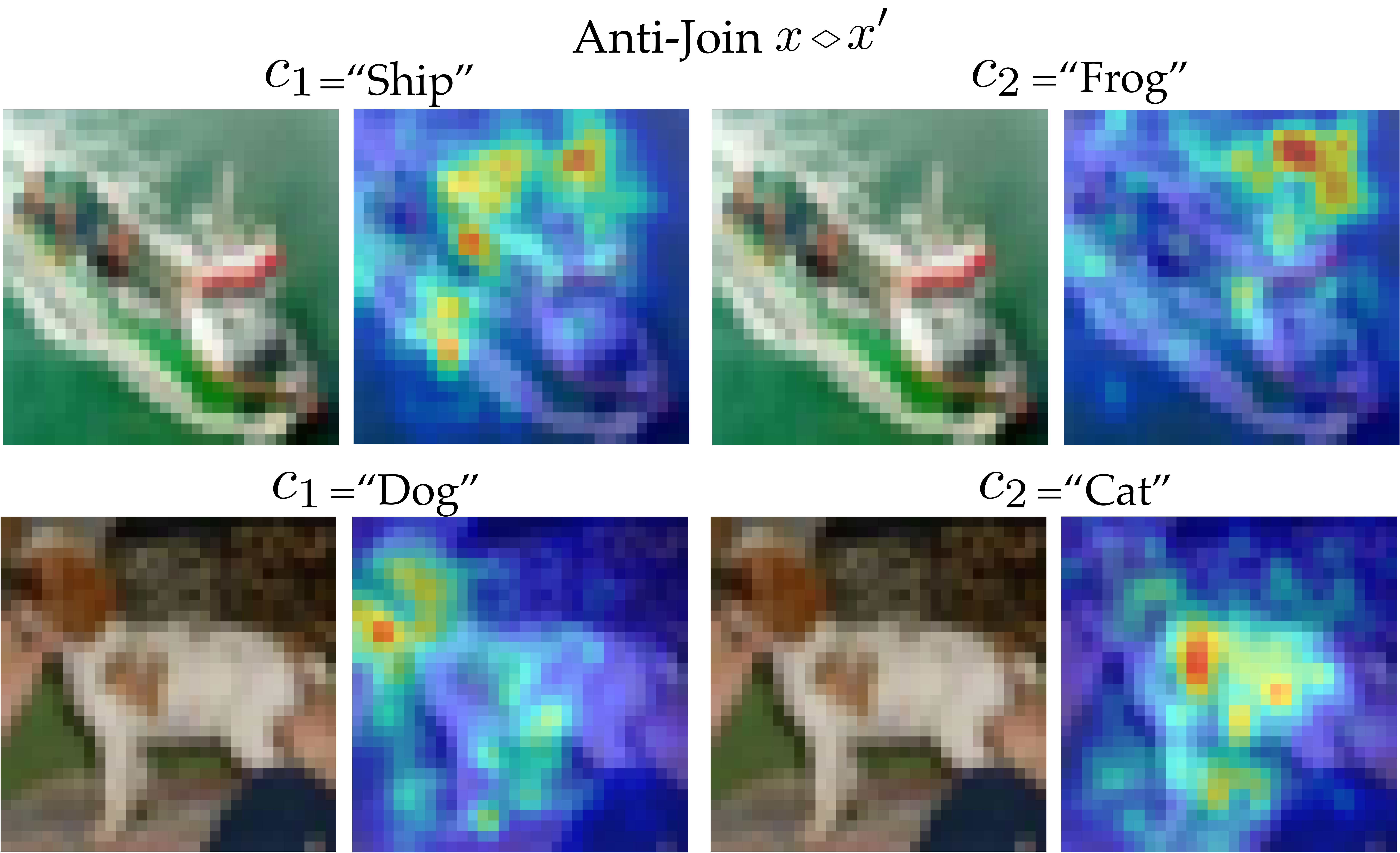}
\caption{Sample inputs and their interpretation under the anti-join operator (CIFAR10 and VGG19). \label{fig:antijoin}}
\end{figure}

Figure\mref{fig:antijoin} compares sample inputs and their attribution maps under the join operator. In each pair, one is a legitimate input and classified correctly (e.g., ``ship'' and ``dog''); the other is an adversarial input (crafted by the PGD attack\mcite{madry:iclr:2018}) and misclassified (e.g., ``frog'' and ``cat''). The anti-join operator highlights the most discriminative features that result in their different predictions.

Note that the anti-join operator is extensible to the case of the same input $x$ but different models $f$ and $f'$. Specifically, to compute $x$'s features that discriminate $f(x)$ from $f'(x)$, we update the expectation in \meq{eq:basic} as $\ssub{\mathbb{E}}{I_k}[f(\ssub{x}{I_k \cup \{i\} }) - f'( \ssub{x}{I_k})]$. Intuitively, for $i$-th feature, we compute the difference of its contribution with respect to $f(x)$ and $f'(x)$.

\subsection{Compositions}
\label{sec:comp}

The library of atomic operators is naturally {\em composable}. For instance, one may combine the selection and projection operators, $\Pi_w (\sigma_l(x))$, which extracts the interpretation at the $l$-th layer of the DNN and magnifies the features within the window $w$; it is possible to compose the join and selection operators, $\sigma_l(x) \bowtie \sigma_l(x') $, which highlights the most discriminative features of $x$ and $x'$ resulting in their different predictions from the view of the $l$-th layer of the DNN $f$; further, it is possible to combine two anti-join operators, $x_1 \ajoin x_2 \ajoin x_3$, which generates the most discriminative features of each input with respect to the rest two.

To ensure their semantic correctness, one may specify that the compositions of different operators to satisfy certain properties (e.g., commutative). For instance, the composition of the selection ($\sigma$) and projection ($\Pi$) operators needs to satisfy the commutative property, that is, the order of applying the two operators should not affect the interpretation result, $\Pi_{w}\sigma_l(x) = \sigma_l \Pi_w(x)$å. Moreover, certain compositions are allowed only under certain conditions (conditional). For instance, the composition of two selection operators, $\sigma_l\sigma_{l'}(x)$, is only defined if $l \leq l'$, that is, it generates the interpretation of $x$ with respect to the layer with the smaller index in $l$ and $l'$ of the DNN $f$. Further, the composition of the join and anti-join operators are undefined. 

%% file: analysis.tex
\section{Interactive Interpretation}
\label{sec:analysis}

Further, \system offers a declarative language that allows users to easily ``query'' the interpretation of DNN behaviors and build interactive analysis tasks as combinations of queries (cf. Figure\mref{fig:stack}).

\subsection{A Declarative Query Language}

Specifically, we define an SQL-like declarative query language for interpreting DNN behaviors. Next we first define the statements for each atomic operator and then discuss their compositions.

We use a set of keywords: {\sf ``select''} for the selection operator, {\sf ``where''} for the projection operator, {\sf ``join''} for the join operator, and {\sf ``left join''} for the anti-join operator. The atomic operators can be invoked using the following statements:

\begin{minted}[mathescape,
escapeinside=||,
fontsize=\scriptsize,
]{sql}
select * from f(x)
\end{minted}
-- the identity operator $\phi(x)$.

\begin{minted}[mathescape,
escapeinside=||,
fontsize=\scriptsize,
]{sql}
select * from f(x) where w
\end{minted}
-- the projection operator $\Pi_w(x)$.

\begin{minted}[mathescape,
escapeinside=||,
fontsize=\scriptsize,
]{sql}
select l from f(x)
\end{minted}
-- the selection operator $\sigma_l(x)$.

\begin{minted}[mathescape,
escapeinside=||,
fontsize=\scriptsize,
]{sql}
select * from f(x) join (select * from f(x|'|))
\end{minted}
-- the join operator $x \bowtie x'$.

\begin{minted}[mathescape,
escapeinside=||,
fontsize=\scriptsize,
]{sql}
select * from f(x) left join (select * from f(x|'|))
\end{minted}
-- the anti-join operator $x \ajoin x'$.

Similar to the concept of ``sub-queries'' in SQL, more complicated operators can be built by composing the statements of atomic operators. Following are a few examples.

\begin{minted}[mathescape,
escapeinside=||,
fontsize=\scriptsize,
]{sql}
select l from f(x) where w
\end{minted}
-- the composition of selection and projection $\Pi_w\sigma_l(x)$.

\begin{minted}[mathescape,
escapeinside=||,
fontsize=\scriptsize,
]{sql}
select l from f(x) join (select l from f(x|'|))
\end{minted}
-- the composition of join and selection $\sigma_l(x)\bowtie \sigma_l(x')$.

\subsection{Interactive Analysis}

Through the declarative queries, users are able to conduct an in-depth analysis of DNN behaviors, including:

\vspace{1pt}
\textbf{Drill-Down Analysis --} Here the user applies a sequence of projection and/or selection to investigate how the \dnn model $f$ classifies a given input $x$ at different granularities of $x$ and at different stages of $f$. This analysis helps answer important questions such as: (i) how does the importance of $x$'s features evolve through different stages of $f$? (ii) which parts of $x$ are likely to be the cause of its misclassification? (iii) which stages of $f$ do not function as expected?

\vspace{1pt}
\textbf{Comparative Analysis --} In a comparative analysis, the user applies a combination of join and/or anti-join operators on the target input $x$ and a set of reference inputs $\mathcal{X}$ to compare how the \dnn $f$ processes $x$ and $x' \in \mathcal{X}$. This analysis helps answer important questions, including: (i) from $f$'s view, why are $x$ and $x' \in \mathcal{X}$ similar or different? (ii) does $f$ indeed find the discriminative features of $x$ and $x'\in \mathcal{X}$? (iii) if $x$ is misclassified into the class of $x'$, which parts of $x$ are likely to be the cause?

\vspace{1pt}
\textbf{What-If Analysis --} In what-if analysis, the user modifies parts of the input $x$ before applying the operators and compares the interpretation before and after the modification. The modification may include (i) nullification (e.g., replacing parts of $x$ with baseline), (ii) substitution (e.g., substituting parts of $x$ with another input), and (iii) transformation (e.g., scaling, rotating, shifting). This analysis allows the user to analyze $f$'s sensitivity to each part of $x$ and its robustness against perturbation.

Note that these tasks are not exclusive; rather, they may complement each other by providing different perspectives on the behaviors of \dnn models. For instance, both drill-down and what-if analyses help the user gauge the impact of $x$'s part $x[w]$ on $f$'s prediction; yet, the former focuses on analyzing the decision boundary within the space spanned by the features $w$, while the latter focuses on analyzing the overall contribution of $x[w]$ to $f$'s prediction.

%% file: evaluation.tex
\section{Empirical Evaluation}
\label{sec:eval}

We prototype \system and empirically evaluate its usability in a set of case studies. The evaluation is designed to answer the following key questions.
\begin{mitemize}
\setlength\itemsep{0.5pt}
\item RQ1: Versatility -- Does \system effectively support a range of analysis tasks?
\item RQ2: Effectiveness -- Does it significantly improve the analysis efficacy in such tasks?
\item RQ3: Usability -- Does it provide intuitive, user-friendly interfaces for analysts?
\end{mitemize}

We conduct user studies on the Amazon Mechanical Turk platform, in which each task involves 1,250 assignments conducted by 50 qualified workers. We apply the following quality control: (i) the users are instructed about the task goals and declarative queries,
and (ii) the tasks are set as small batches to reduce bias and exhaustion.



\subsection{Case A: Resolving Model Inconsistency}

Two DNNs trained for the same task often differ slightly due to (i) different training datasets, (ii) different training regimes, and (iii) randomness inherent in training algorithms (e.g., random shuffle and dropout). It is thus critical to identify the correct one when multiple DNNs disagree on the prediction on a given input. In this case study, the user is requested to use \system to resolve cases that are inconsistent between two DNNs $f$ and $f'$.

\textbf{Setting --} On CIFAR10, we train two VGG19 models $f$ and $f'$. In the testing set of CIFAR10, 946 samples are predicted differently by $f$ and $f'$, in which 261 samples are correctly predicted by $f$ and 565 samples by $f'$. Within this set, 824 inputs are classified correctly by either $f$ or $f'$, which we collect as the testing set $\gT$ for our study.

\begin{figure}[!ht]
\centering
\includegraphics[width=60mm]{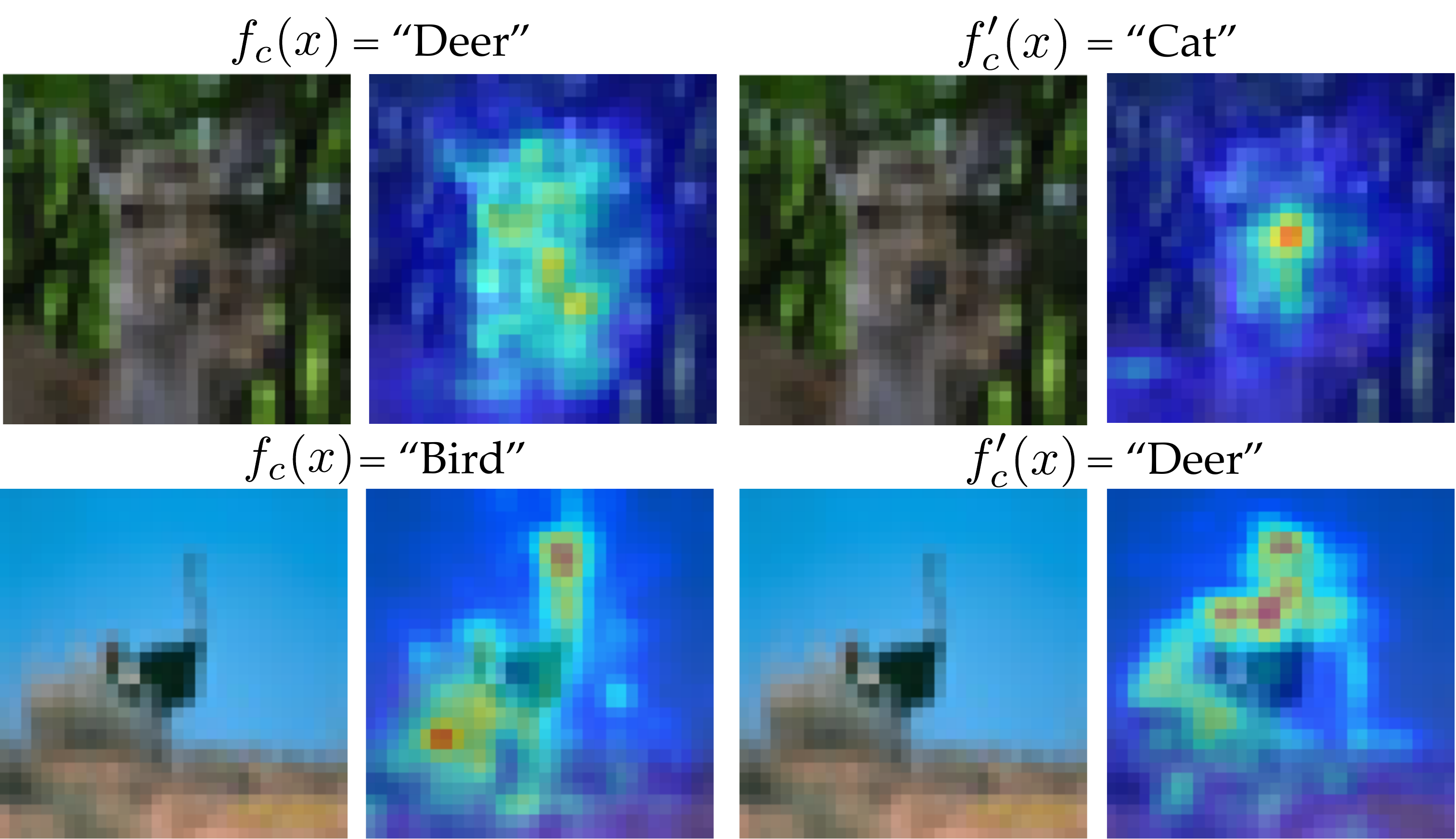}
\caption{Sample inputs, their classification by $f$ and $f'$, and their interpretation by \system in Case A. \label{fig:mod_samples}}
\end{figure}

We randomly sample 50 (half predicted correctly by $f$ and the other half by $f'$) inputs from $\gT$ to form the testing set. Specifically, the baseline directly generates the interpretation $f(x)$ and $f'(x)$ for each input $x$;
\system applies the \emph{Anti-Join} operator to highlight $x$'s most discriminative features (from the views of $f$ and $f'$) that result in its different predictions, with the declarative query given as: \mintinline{sql}{select * from f(x) left join (select * from f'(x))}. Figure\mref{fig:mod_samples} shows a set of sample inputs, their classification under $f$ and $f'$, and their interpretation by \system. Observe that the discriminative features on the correct model tend to agree with human perception better.

\textbf{Evaluation --} We evaluate \system in terms of (i) effectiveness -- whether it helps users identify the correct predictions, and (ii) efficiency -- whether it helps users conduct the analysis more efficiently. We measure the effectiveness using the metric of accuracy (the fraction of correctly distinguished inputs among total inputs) and assess the efficiency using the metric of user response time (URT), which is measured by the average time the users spend on each input.

\begin{figure}[!ht]
\centering
\includegraphics[width=85mm]{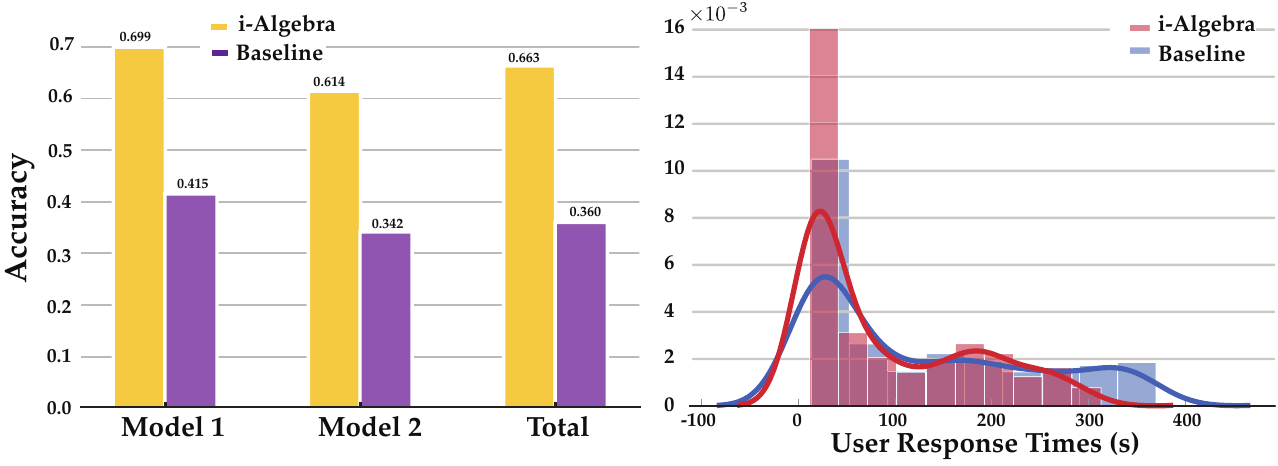}
\caption{Users' accuracy and URT measures under baseline and \system in Case A. \label{fig:mod_combine}}
\end{figure}

Figure\mref{fig:mod_combine} compares the users' performance using the baseline and \system on the task of resolving model inconsistency. We have the following observations: (i) \system significantly improves the users' accuracy of identifying the correct predictions on both $f$ and $f'$, with the overall accuracy increasing by around 30\%; (ii) Despite its slightly more complicated interfaces, the URT on \system does not observe a significant change from the baseline, highlighting its easy-to-use mechanisms and interfaces.

\subsection{Case B: Detecting Adversarial Inputs}

One intriguing property of \dnns is their vulnerability to adversarial inputs, which are maliciously crafted samples to deceive target DNNs\mcite{madry:iclr:2018, carlini:sp:2017}. Adversarial inputs are often generated by carefully perturbing benign inputs, with difference imperceptible to human perception.
Recent work has proposed to leverage interpretation as a defense mechanism to detect adversarial inputs\mcite{tao:nips:2018}. Yet, it is shown that the interpretation model is often misaligned with the underlying DNN model, resulting in the possibility for the adversary to deceive both models simultaneously\mcite{Zhang:2020:sec}.
In this use case, the users are requested to leverage \system to inspect potential adversarial inputs from multiple different interpretation perspectives, making it challenging for the adversary to evade the detection across all such views.

\vspace{1pt}
\textbf{Setting --} We use ImageNet as the dataset and consider a pre-trained ResNet50 (77.15\% top-1 accuracy) as the target DNN. We also train a set of truncated DNN models ($l=2, 3, 4$) for the selection operator $\sigma_l(x)$ in \system (details in \msec{sec:algebra}). We apply \textsf{ADV}$^2$\mcite{Zhang:2020:sec}, an attack designed to generate adversarial inputs deceiving both the DNN and its coupled interpreter. Specifically, \textsf{ADV}$^2$ optimizes the objective function:
\begin{equation}
\begin{array}{cl}
{\min _{x}} & {\ssub{\ell}{\text{prd}}\left(f(x), \ssub{c}{t}\right)+\lambda \ssub{\ell}{\text{int}}\left(g(x ; f), \ssub{m}{t}\right)} \\
{\text { s.t. }} & {\Delta\left(x, \bx\right) \leq \epsilon}
\end{array}
\end{equation}
where $\ssub{\ell}{\text{prd}}$ ensures that the adversarial input $x$ is misclassified to a target class $\ssub{c}{t}$ by the DNN $f$, and $\ssub{\ell}{\text{int}}$ ensures that $x$ generates an attribution map similar to a target map $\ssub{m}{t}$ (the attribution map of the benign input $\bx$).

\begin{figure}[ht]
\centering
\includegraphics[width=60mm]{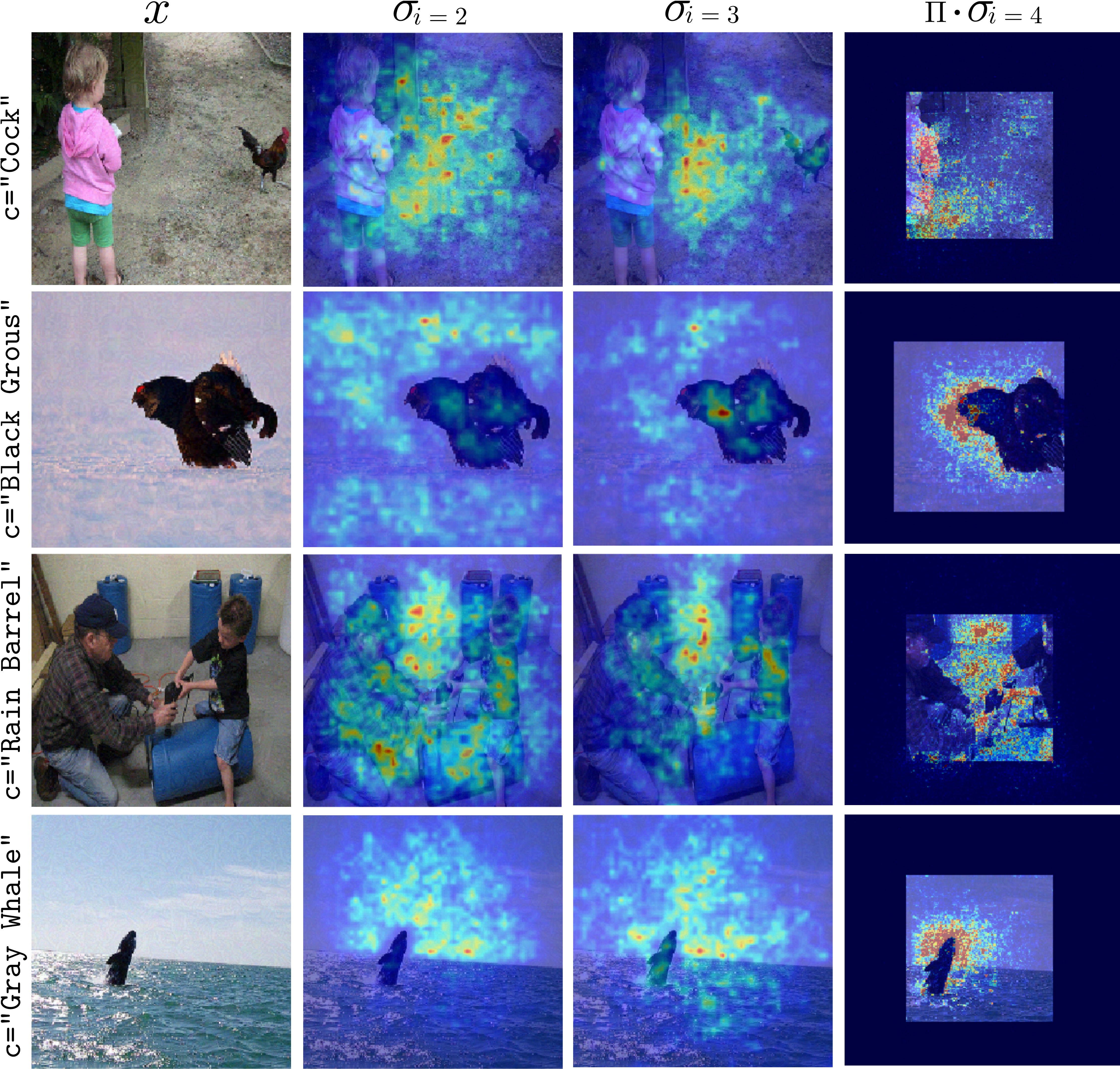}
\caption{Sample adversarial inputs and \system interpretation.
\label{fig:adv-all}}
\end{figure}

From ImageNet, we randomly sample 50 inputs and generate their adversarial counterparts, which are combined with another 50 randomly sampled benign inputs to form the testing set $\gT$. We request the users to identify the adversarial inputs through the lens of baseline and \system. In particular, by varying $l$ and $w$, \system provides interpretation at various inference stages and input granularity, with the declarative query template as:
\mintinline{sql}{select l from f(x) where w}. Figure\mref{fig:adv-all} shows sample adversarial inputs and their interpretation under \system. Observe that by from multiple complementary perspectives, the adversarial inputs show fairly distinguishable interpretation.

\vspace{1pt}
\textbf{Evaluation --}
We quantitatively assess the usability of \system in terms of (i) effectiveness -- whether it helps users identify the adversarial inputs more accurately, and (ii) efficiency -- whether it helps users conduct the analysis more efficiently.
Specifically, considering adversarial and benign inputs as positive and negative cases, we measure the effectiveness using the metrics of precision and recall (as well as accuracy and F-1 score):

We assess the efficiency using the metric of user response time (URT), which is measured by the average time the users spend on each task with the given tool.

\begin{figure}[!ht]
\centering
\epsfig{file = 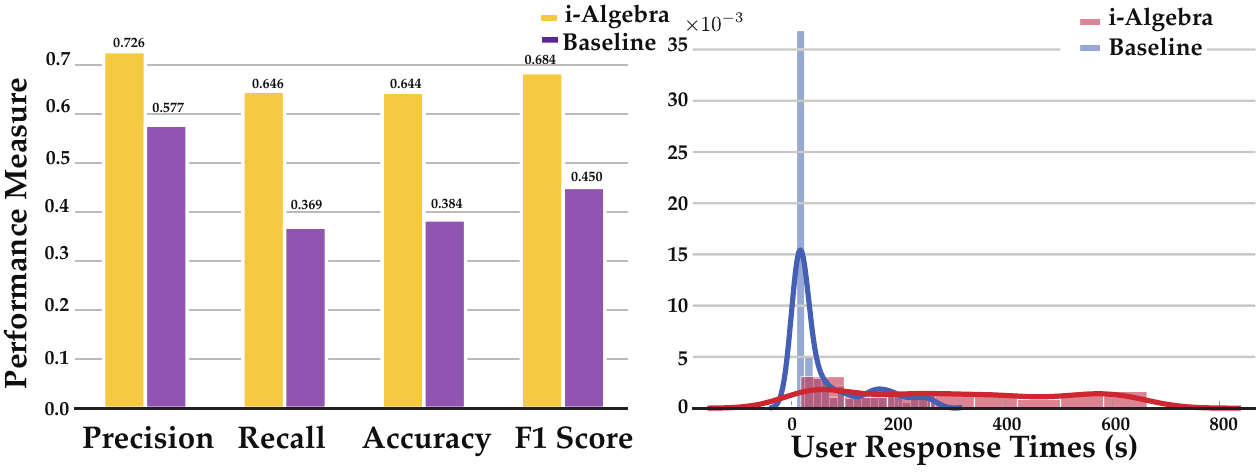, width=85mm}
\caption{Users' performance and URT measures under baseline and \system in Case B. \label{fig:adv-combine}}
\end{figure}

Figure\mref{fig:adv-combine} compares the users' performance and URT using the baseline interpretation and \system on the task of identifying adversarial inputs. We have the following observations: (i) through the lens of interpretation from multiple complementary perspectives, \system improves the users' effectiveness of distinguishing adversarial and benign inputs with about 26\% increase in the overall accuracy; (ii) compared with the baseline, the average URT on \system grows from 60.45s to 303.54s, which can be intuitively explained by its requirement for multiple rounds of interactive queries. Given the significant performance improvement, the cost of execution time is well justified.

\begin{figure}[!ht]
\centering
\includegraphics[width=60mm]{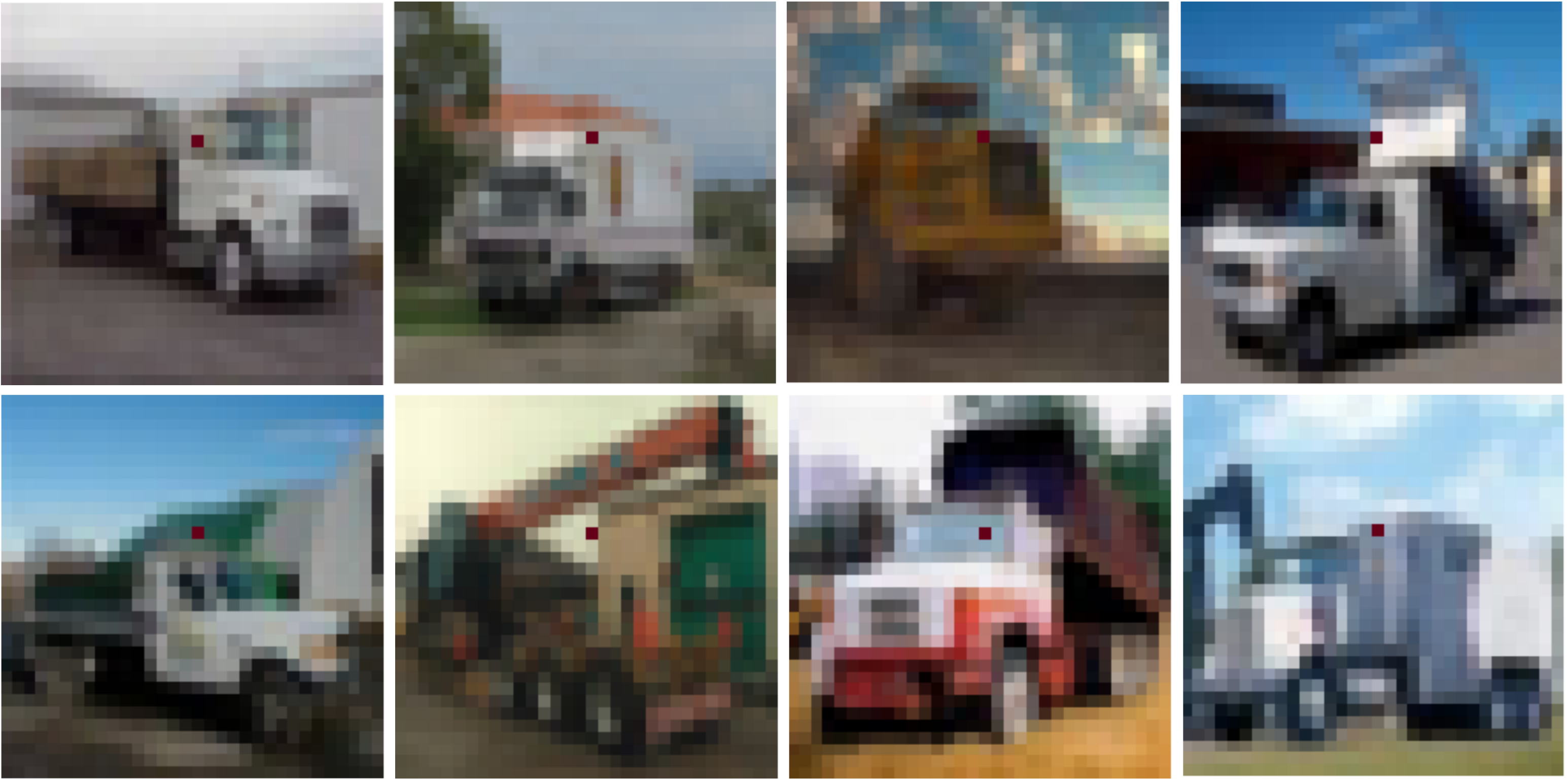}
\caption{Sample trigger-embedded inputs at inference, which are misclassified as ``deer''. \label{fig:poi_examples}}
\end{figure}

\subsection{Case C: Cleansing Poisoning Data}

Orthogonal to adversarial examples, another concern for the safety of DNNs is their vulnerability to manipulations of their training data. In the backdoor attacks (e.g., \mcite{Gu:arxiv:2017}), by injecting corrupted inputs in training a DNN, the adversary forces the resultant model to (i) misclassify the inputs embedded with particular patterns (``triggers'') to a target class and (ii) behave normally on benign inputs. Figure\mref{fig:poi_examples} shows a set of trigger-embedded inputs which are misclassified from ``truck'' to ``deer''.

Since the backdoored DNNs correctly classify benign inputs, once trained, they are insidious to be detected. One mitigation is to detect corrupted instances in the training set, and then to use cleansed data to re-train the DNN\mcite{spectral:2018:nips}. Typically, the analyst applies statistical analysis on the deep representations of the training data and detects poisoning inputs based on their statistical anomaly.

We let the analyst leverage \system to perform fine-tuning of the results detected by the statistical analysis. Through the lens of the interpretation, the users may inspect the inputs that fall in the uncertain regions (e.g., 1.25 $\sim$ 1.75 standard deviation) and identify false positives and false negatives by the automated detection, which may further improve the mitigation of backdoor attacks.

\begin{table}[!ht]
\centering
{\small
\begin{tabular}{c|c|c}
\multirow{2}{*}{Prediction} & \multicolumn{2}{c}{Ground-truth}\\
\cline{2-3}
& + & - \\ \cline{2-3}
\hline
\hline
+ & 48 & 654 \\
\hline
- & 399 & 3574 \\
\hline
\end{tabular}}
\caption{Samples statistics in Case C.
\label{tab:usecase3-truetable}}
\end{table}

\vspace{1pt}
\textbf{Setting --} We use CIFAR10 as the underlying dataset and VGG19 as the target DNN. We consider using the backdoor attack in\mcite{Gu:arxiv:2017} to generate poisoning instances in one particular class ``truck''. We apply spectral signature\mcite{spectral:2018:nips} to identify potential poisoning instances. For each data point $x_i$ in a particular class, it examines $x_i$'s deep representation $\mathcal{R} (x_{i})$ at the penultimate layer. After obtaining the top right singular vector of the centered representation $[ \mathcal{R} (x_{i} )- \frac{1}{n} \sum_{i=1}^{n} \mathcal{R} (x_{i}) ]_{i=1}^{n}$, the data points beyond 1.5 standard deviation from the center are identified as poisonous. However, this automated tool is fairly inaccurate. Table\mref{tab:usecase3-truetable} summarizes the predicted results and the ground truth. In this case, we request users to identify positive and negative cases that are misclassified by the automated tool, with the declarative query template given as: \mintinline{sql}{select l from f(x)}.

\begin{table}[!ht]
\centering
{\small
\begin{tabular}{c|c|c|c}
\multicolumn{1}{c|}{Precision} & \multicolumn{1}{c|}{Recall} & \multicolumn{1}{c|}{Accuracy} & \multicolumn{1}{c}{ F1-Score} \\ \cline{2-3}
\hline
\hline
0.609 & 0.6343& 0.586 & 0.622 \\
\hline
\end{tabular}}
\caption{Users' performance under \system in Case C.
\label{tab:usecasec-stat}}
\end{table}

\begin{figure}[!ht]
\centering
\includegraphics[width=50mm]{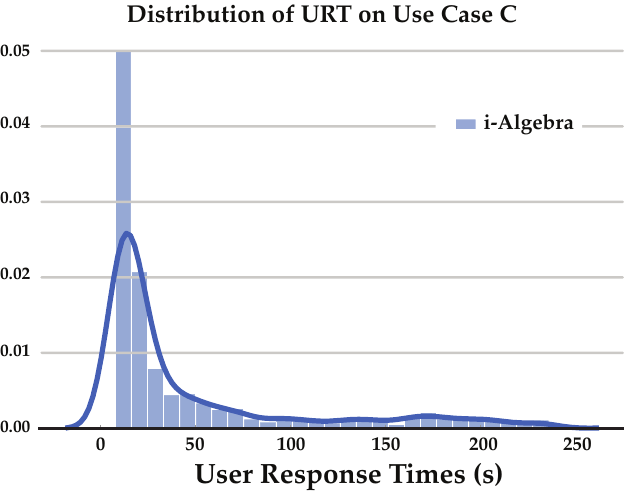}
\caption{URT distribution in Case C. \label{fig:usecasec}}
\end{figure}

\vspace{1pt}
\textbf{Evaluation --} We measure the users' effectiveness of distinguishing positive and negative cases. The results are listed in Table\mref{tab:usecasec-stat}. Observe that equipped with \system, the users successfully classify around 60\% of the data point misclassified by the automated tool. Figure \mref{fig:usecasec} shows the distribution of URT for this task. Note that a majority of users take less than 50s to complete the tasks.

%% file: literature.tex
\section{Related work}
\label{sec:literature}

Next, we survey three categories of prior work: interpretable deep learning, model attacks and defenses, and interactive learning.

\vspace{1pt}
\textbf{Interpretable Deep Learning --}
Typically, DNN interpretability can be obtained by either designing interpretable models\mcite{Zhang:2018:cvpr} or extracting post-hoc interpretations. The post-hoc interpretation methods can be categorized as backprop-\mcite{Sundararajan:2017:icml}, representation-\mcite{selvaraju:gradcam}, meta-model-\mcite{Dabkowski:nips:2017}, and perturbation-based\mcite{fong:mask}. Instead of developing yet another interpretation method or enhancing existing ones, this work proposes the paradigm of interactive interpretability, which can be flexibly implemented upon existing methods.

\vspace{1pt}
\textbf{Model Attacks and Defenses --}
DNNs are becoming the new targets of malicious attacks, including adversarial attacks\mcite{carlini:sp:2017} and poisoning attacks\mcite{Shafahi:2018:nips}. Although a line of work strives to improve \dnn robustness\mcite{Tramer:2018:iclr,spectral:2018:nips}, existing defenses are often penetrated by even stronger attacks\mcite{Ling:2019:sp}, resulting in a constant arms race.
Our work involves the users in the process of model robustness improvement, which is conducive to enhancing model trustworthiness.

\vspace{1pt}
\textbf{ Interactive Learning --}
Interactive learning couples humans and machine learning models tightly within the learning process. The existing work can be roughly categorized as model understanding\mcite{krause:2016:icml,krause:2016:chi}, which allows users to interpret models' input-output dependence, and model debugging\mcite{errudite:2019:acl,accountableai:2018:aaai}, which allows users to detect and fix models' mistakes.
\system can be leveraged to not only understand \dnns' behaviors but also facilitate diverse security tasks including model debugging, data cleansing, and attack inspection.

%% file: conclusion.tex
\section{Conclusion}
\label{sec:conclusion}

This work promotes a paradigm shift from {\em static} interpretation to {\em interactive} interpretation of neural networks, which we believe will significantly improve the usability of existing interpretation models in practice. We present \system, a first-of-its-kind interactive framework for DNN interpretation. At its core is a library of atomic operators that produce the interpretation of DNN behaviors at varying input granularity, at different inference stages, and from distinct interpretation perspectives. A declarative query language is defined for users to flexibly construct a variety of analysis tasks by composing different operators. We prototype \system and conduct extensive studies in three representative analysis tasks, all demonstrating its promising usability.

%% file: main.bbl
\begin{thebibliography}{23}
\providecommand{\natexlab}[1]{#1}
\providecommand{\url}[1]{\texttt{#1}}
\providecommand{\urlprefix}{URL }
\expandafter\ifx\csname urlstyle\endcsname\relax
  \providecommand{\doi}[1]{doi:\discretionary{}{}{}#1}\else
  \providecommand{\doi}{doi:\discretionary{}{}{}\begingroup
  \urlstyle{rm}\Url}\fi

\bibitem[{{Ancona}, {{\"O}ztireli}, and {Gross}(2019)}]{Ancona:2019:icml}
{Ancona}, M.; {{\"O}ztireli}, C.; and {Gross}, M. 2019.
\newblock {Explaining Deep Neural Networks with a Polynomial Time Algorithm for
  Shapley Values Approximation}.
\newblock In \emph{Proceedings of IEEE Conference on Machine Learning (ICML)}.

\bibitem[{Carlini and Wagner(2017)}]{carlini:sp:2017}
Carlini, N.; and Wagner, D.~A. 2017.
\newblock {Towards Evaluating the Robustness of Neural Networks}.
\newblock In \emph{Proceedings of IEEE Symposium on Security and Privacy
  (S\&P)}.

\bibitem[{{Chen} et~al.(2019){Chen}, {Song}, {Wainwright}, and
  {Jordan}}]{Chen:2019a:iclr}
{Chen}, J.; {Song}, L.; {Wainwright}, M.~J.; and {Jordan}, M.~I. 2019.
\newblock {L-Shapley and C-Shapley: Efficient Model Interpretation for
  Structured Data}.
\newblock In \emph{Proceedings of International Conference on Learning
  Representations (ICLR)}.

\bibitem[{{Dabkowski} and {Gal}(2017)}]{Dabkowski:nips:2017}
{Dabkowski}, P.; and {Gal}, Y. 2017.
\newblock {Real Time Image Saliency for Black Box Classifiers}.
\newblock In \emph{Proceedings of Advances in Neural Information Processing
  Systems (NeurIPS)}.

\bibitem[{Fong and Vedaldi(2017)}]{fong:mask}
Fong, R.~C.; and Vedaldi, A. 2017.
\newblock {Interpretable Explanations of Black Boxes by Meaningful
  Perturbation}.
\newblock In \emph{Proceedings of IEEE International Conference on Computer
  Vision (ICCV)}.

\bibitem[{Goodfellow, Bengio, and Courville(2016)}]{Goodfellow-et-al-2016}
Goodfellow, I.; Bengio, Y.; and Courville, A. 2016.
\newblock \emph{{Deep Learning}}.
\newblock MIT Press.

\bibitem[{{Gu}, {Dolan-Gavitt}, and {Garg}(2017)}]{Gu:arxiv:2017}
{Gu}, T.; {Dolan-Gavitt}, B.; and {Garg}, S. 2017.
\newblock {BadNets: Identifying Vulnerabilities in the Machine Learning Model
  Supply Chain}.
\newblock \emph{ArXiv e-prints} .

\bibitem[{Guo et~al.(2018)Guo, Mu, Xu, Su, Wang, and Xing}]{Guo:2018:ccs}
Guo, W.; Mu, D.; Xu, J.; Su, P.; Wang, G.; and Xing, X. 2018.
\newblock {LEMNA: Explaining Deep Learning Based Security Applications}.
\newblock In \emph{Proceedings of ACM SAC Conference on Computer and
  Communications (CCS)}.

\bibitem[{{Karpathy}, {Johnson}, and {Fei-Fei}(2016)}]{Karpathy:2016:iclr}
{Karpathy}, A.; {Johnson}, J.; and {Fei-Fei}, L. 2016.
\newblock {Visualizing and Understanding Recurrent Networks}.
\newblock In \emph{Proceedings of International Conference on Learning
  Representations (ICLR)}.

\bibitem[{Krause, Perer, and Bertini(2016)}]{krause:2016:icml}
Krause, J.; Perer, A.; and Bertini, E. 2016.
\newblock {Using Visual Analytics to Interpret Predictive Machine Learning
  Models}.
\newblock In \emph{Proceedings of IEEE Conference on Machine Learning (ICML)}.

\bibitem[{Krause, Perer, and Ng(2016)}]{krause:2016:chi}
Krause, J.; Perer, A.; and Ng, K. 2016.
\newblock {Interacting with Predictions: Visual Inspection of Black-Box Machine
  Learning Models}.
\newblock In \emph{Proceedings of the CHI Conference on Human Factors in
  Computing Systems (CHI)}.

\bibitem[{Ling et~al.(2019)Ling, Ji, Zou, Wang, Wu, Li, and
  Wang}]{Ling:2019:sp}
Ling, X.; Ji, S.; Zou, J.; Wang, J.; Wu, C.; Li, B.; and Wang, T. 2019.
\newblock {DEEPSEC: A Uniform Platform for Security Analysis of Deep Learning
  Model}.
\newblock In \emph{Proceedings of IEEE Symposium on Security and Privacy
  (S\&P)}.

\bibitem[{Madry et~al.(2018)Madry, Makelov, Schmidt, Tsipras, and
  Vladu}]{madry:iclr:2018}
Madry, A.; Makelov, A.; Schmidt, L.; Tsipras, D.; and Vladu, A. 2018.
\newblock {Towards Deep Learning Models Resistant to Adversarial Attacks}.
\newblock In \emph{Proceedings of International Conference on Learning
  Representations (ICLR)}.

\bibitem[{Nushi, Kamar, and Horvitz(2018)}]{accountableai:2018:aaai}
Nushi, B.; Kamar, E.; and Horvitz, E. 2018.
\newblock {Towards Accountable AI: Hybrid Human-Machine Analyses for
  Characterizing System Failure}.
\newblock In \emph{Proceedings of AAAI Conference on Artificial Intelligence
  (AAAI)}.

\bibitem[{Selvaraju et~al.(2017)Selvaraju, Cogswell, Das, Vedantam, Parikh, and
  Batra}]{selvaraju:gradcam}
Selvaraju, R.~R.; Cogswell, M.; Das, A.; Vedantam, R.; Parikh, D.; and Batra,
  D. 2017.
\newblock {Grad-CAM: Visual Explanations from Deep Networks via Gradient-Based
  Localization}.
\newblock In \emph{Proceedings of IEEE International Conference on Computer
  Vision (ICCV)}.

\bibitem[{{Shafahi} et~al.(2018){Shafahi}, {Ronny Huang}, {Najibi}, {Suciu},
  {Studer}, {Dumitras}, and {Goldstein}}]{Shafahi:2018:nips}
{Shafahi}, A.; {Ronny Huang}, W.; {Najibi}, M.; {Suciu}, O.; {Studer}, C.;
  {Dumitras}, T.; and {Goldstein}, T. 2018.
\newblock {Poison Frogs! Targeted Clean-Label Poisoning Attacks on Neural
  Networks}.
\newblock In \emph{Proceedings of Advances in Neural Information Processing
  Systems (NeurIPS)}.

\bibitem[{Sundararajan, Taly, and Yan(2017)}]{Sundararajan:2017:icml}
Sundararajan, M.; Taly, A.; and Yan, Q. 2017.
\newblock {Axiomatic Attribution for Deep Networks}.
\newblock In \emph{Proceedings of IEEE Conference on Machine Learning (ICML)}.

\bibitem[{Tao et~al.(2018)Tao, Ma, Liu, and Zhang}]{tao:nips:2018}
Tao, G.; Ma, S.; Liu, Y.; and Zhang, X. 2018.
\newblock {Attacks Meet Interpretability: Attribute-Steered Detection of
  Adversarial Samples}.
\newblock In \emph{Proceedings of Advances in Neural Information Processing
  Systems (NeurIPS)}.

\bibitem[{{Tram{\`e}r} et~al.(2018){Tram{\`e}r}, {Kurakin}, {Papernot},
  {Goodfellow}, {Boneh}, and {McDaniel}}]{Tramer:2018:iclr}
{Tram{\`e}r}, F.; {Kurakin}, A.; {Papernot}, N.; {Goodfellow}, I.; {Boneh}, D.;
  and {McDaniel}, P. 2018.
\newblock {Ensemble Adversarial Training: Attacks and Defenses}.
\newblock In \emph{Proceedings of International Conference on Learning
  Representations (ICLR)}.

\bibitem[{Tran, Li, and Madry(2018)}]{spectral:2018:nips}
Tran, B.; Li, J.; and Madry, A. 2018.
\newblock {Spectral Signatures in Backdoor Attacks}.
\newblock In \emph{Proceedings of Advances in Neural Information Processing
  Systems (NeurIPS)}.

\bibitem[{Wu et~al.(2019)Wu, Ribeiro, Heer, and Weld}]{errudite:2019:acl}
Wu, T.; Ribeiro, M.~T.; Heer, J.; and Weld, D.~S. 2019.
\newblock {Errudite: Scalable, Reproducible, and Testable Error Analysis}.
\newblock In \emph{Proceedings of Annual Meeting of the Association for
  Computational Linguistics (ACL)}.

\bibitem[{{Zhang}, {Nian Wu}, and {Zhu}(2018)}]{Zhang:2018:cvpr}
{Zhang}, Q.; {Nian Wu}, Y.; and {Zhu}, S.-C. 2018.
\newblock {Interpretable Convolutional Neural Networks}.
\newblock In \emph{Proceedings of IEEE Conference on Computer Vision and
  Pattern Recognition (CVPR)}.

\bibitem[{{Zhang} et~al.(2020){Zhang}, {Wang}, {Shen}, {Ji}, {Luo}, and
  {Wang}}]{Zhang:2020:sec}
{Zhang}, X.; {Wang}, N.; {Shen}, H.; {Ji}, S.; {Luo}, X.; and {Wang}, T. 2020.
\newblock {Interpretable Deep Learning under Fire}.
\newblock In \emph{Proceedings of USENIX Security Symposium (SEC)}.

\end{thebibliography}
